%% file: main.tex
\documentclass[runningheads]{llncs}
\usepackage[T1]{fontenc}
\usepackage{graphicx}
\usepackage{float}
\usepackage{algorithm}
\usepackage[noend]{algpseudocode}
\usepackage{wrapfig,lipsum,booktabs}
\usepackage{booktabs} 
\usepackage{enumitem}

\usepackage{graphicx}

\usepackage{hyperref}
\usepackage{cite}

\usepackage{amsmath}
\usepackage{amssymb}
\usepackage{mathtools}

\begin{document}
\title{Data Augmentation for Continual RL via Adversarial Gradient Episodic Memory}
%
\titlerunning{Data Augmentation for Continual RL via Adv-GEM}
%
\author{Sihao Wu\inst{1}  \and
Xingyu Zhao\inst{2} \and Xiaowei Huang\inst{1}}
\authorrunning{S. Wu et al.}
%
\institute{University of Liverpool, Liverpool, UK\\
\email{sihao.wu@liverpool.ac.uk, xiaowei.huang@liverpool.ac.uk}
\and
WMG, University of Warwick, Warwick, UK\\
\email{
xingyu.zhao@warwick.ac.uk}}

\maketitle              
\begin{abstract}
Data efficiency of learning, which plays a key role in the Reinforcement Learning (RL) training process, becomes even more important in continual RL with sequential environments. 
In continual RL, the learner interacts with non-stationary, sequential tasks and is required to learn new tasks without forgetting previous knowledge. 
However, there is little work on implementing data augmentation for continual RL. In this paper, we investigate the efficacy of data augmentation for continual RL. Specifically, we provide benchmarking data augmentations for continual RL, by (1) summarising existing data augmentation methods and (2) including a new augmentation method for continual RL: Adversarial Augmentation with Gradient Episodic Memory (Adv-GEM). Extensive experiments show that data augmentations, such as random amplitude scaling, state-switch, mixup, adversarial augmentation, and Adv-GEM, can improve existing continual RL algorithms in terms of their average performance, catastrophic forgetting, and forward transfer, on robot control tasks. All data augmentation methods are implemented as plug-in modules for trivial integration into continual RL methods. Our code is available at \url{https://github.com/WilliamWu96/DA_CRL_AGEM}.

\end{abstract}
\input{content/introduction}

\input{content/relatedwork.tex}
\input{content/method}
\input{content/experiment}

\input{content/conclusion}

\noindent \textbf{Acknowledgements.} We thank the anonymous reviewers for their thoughtful comments. XH's contribution is supported by the UK EPSRC through End-to-End Conceptual Guarding of Neural Architectures [EP/T026995/1].


%
%
%
\bibliographystyle{splncs04}
\bibliography{reference}

\end{document}

%% file: content/introduction.tex
\section{Introduction}
\label{submission}

Reinforcement Learning (RL) is widely adapted to various fields \textit{e.g.}, robotics \cite{singh2022reinforcement,fang2017learning}, autonomous vehicles \cite{shalev2016safe} and language model improvement \cite{ouyang2022training,liu2024tiny}, to name a few. It is well-established that the performance of RL agents in \emph{dynamic environments} can be significantly compromised by their \textit{poor data efficiency} and limited \textit{generalization ability} \cite{kirk2021survey}. To address this challenge in vanilla RL, numerous studies such as \cite{kostrikov2020image,2111.09794,laskin2020reinforcement,sinha2022s4rl} have employed data augmentation techniques. Various augmentation methods, including image transformation and interpolation \cite{wang2020improving}, have been adapted to RL. Additionally, \cite{zhang2021generalization} introduces the generation of adversarial data augmentations to enhance generalization capabilities.

However, the challenges faced by RL agents become more pronounced in the context of continual RL, where the investigation into effective solutions remains limited. While data augmentation has shown remarkable success in vanilla RL, its potential in continual RL scenarios has yet to be fully explored. Motivated by these promising results in standard RL settings, we aim to address a critical research question that bridges this gap: 
\begin{quote}
\centering
\textit{Does data augmentation improve continual RL? }
\end{quote}

In continual RL where the learner interacts with non-stationary, sequential tasks, it is expected that the learner can learn new knowledge without forgetting that already learned. When confronted with different tasks, the RL agent either struggles to adapt to new tasks or tends to forget previously learned knowledge. To deal with this, a number of continual RL methods have been introduced recently. For instance, regularization-based methods \cite{kirkpatrick2017overcoming,nguyen2018variational,aljundi2018memory,jin2022enhancing} attempt to mitigate forgetting by keeping the model weights close to previous parameters. Modality-based methods prevent the change of learned parameters from previous tasks with hard constraints \cite{mallya2018packnet}. Another category of continual learning, memory-based methods, stores a subset samples \cite{lopez2017gradient,chaudhry2018efficient,guo2020improved,pmlr-v162-wang22v} from previous tasks to tackle the forgetting problem.



We devise an accessible, plug-in module to enhance continual RL methods with data augmentation. That is, the data augmentation framework works without changing the underlying RL algorithm for continual RL. This plug and play nature of our data augmentation scheme enables the extension of arbitrary continual RL methods. 
In terms of the concrete methods, we first summarise several existing data augmentation methods, including Uniform noise, Gaussian noise, random amplitude scaling (RAS), dimension dropout, state-switch, mixup, and adversarial augmentation (Adv-AUG) for state-based continual RL. Then, inspired by Averaged Gradient Episodic Memory (A-GEM) \cite{chaudhry2018efficient}, we introduce a novel augmentation scheme specialized for continual RL, called Adversarial Augmentation with Gradient Episodic Memory (Adv-GEM). The Adv-GEM method indirectly embeds the gradient from previous tasks into the generation of adversarial examples. Finally, to validate the methods, we work with robot control benchmark that has been widely used for robotic manipulation.


In summary, our key contributions include: 
\begin{itemize}[leftmargin=*,label=\textbullet]
\item Pioneering Data Augmentation for Continual RL: This study is the first to comprehensively benchmark data augmentation techniques for state-based continual reinforcement learning. Extensive experiments show that data augmentation methods outperform continual RL baselines in terms of \textit{average performance}, \textit{catastrophic forgetting}, and \textit{forward transfer}. 

\item Enhancing Continual RL Performance: Our experiments reveal that specific data augmentation techniques, particularly RAS, Adv-AUG, and our novel Adv-GEM, substantially improve the average performance of several continual RL algorithms across challenging robotics control benchmarks, including MW4 and CW10. 

\item We introduce a novel data augmentation method, \textbf{Adv-GEM}, specifically designed for continual RL. Our experimental results indicate that Adv-GEM achieves the highest positive impact on average performance compared to other augmentation techniques. 
\end{itemize}

%% file: content/relatedwork.tex
\section{Related Work}
\vspace{-5pt}
\subsection{Continual RL}


In continual RL, considering a sequence of $N$ tasks, the RL agent needs to  handle the induced non-stationarity. A capable continual RL agent should be able to acquire new skills while mitigating catastrophic forgetting. Several works have considered mitigate forgetting by adding a regularization term to keep the model weights \cite{kirkpatrick2017overcoming,nguyen2018variational,aljundi2018memory}. To protect more important weights for previous tasks not changed, Elastic Weight Consolidation (EWC) \cite{kirkpatrick2017overcoming} proposed the Fisher information matrix to estimate the importance of the weights. Another category of method protects any change of the parameters for previous tasks with hard constrain. For instance, PackNet \cite{mallya2018packnet} iterative prunes the model into divided parts for different tasks, freezes the model for the previous task, then retains the model parameters for the current task. Some algorithms keep the memorization by creating episodic memory to store either subset samples \cite{lopez2017gradient,chaudhry2018efficient,guo2020improved,pmlr-v162-wang22v}, or gradient spaces \cite{saha2020gradient} from previous tasks. Within them, A-GEM \cite{chaudhry2018efficient} using the average of the gradients which is calculated by randomly sampling from episodic memory to constrain the gradient direction.

In all, above works focused on the perspective of model modification. However, our method focuses on the perspective of data modification, by adding data augmentation on continual RL, which can be a plug-in module for each continual method mentioned above.

\subsection{Data Augmentation in RL}

Data augmentation has been widely used in the field of NLP \cite{feng2021survey,liang2020mixkd,jindal2020augmenting,zhang2022adversarial} and computer vision \cite{shorten2019survey,cirecsan2011high,ciregan2012multi}. Image transformations, like scales, rotation, translations, color shifts, etc, have been utilized to improve the generalization ability in computer vision. 
This actually inspired the RL community, and recently lots of work came out for the performance increase of RL by adding data augmentation \cite{kostrikov2020image,2111.09794,laskin2020reinforcement,sinha2022s4rl}. While the transformations for image are potentially applicable to image-based RL, \cite{kostrikov2020image,laskin2020reinforcement,raileanu2020automatic,hansen2021stabilizing} actually adapt the augmentation method specifically for RL. Most of the methods, like DrQ \cite{kostrikov2020image}, RAD \cite{laskin2020reinforcement}, UCB-DrAC \cite{raileanu2020automatic}, SVEA \cite{hansen2021stabilizing}, are focused on image-based RL on DeepMind Control suit. In order to improve the sample efficiency, CURL \cite{laskin2020curl} utilized data augmentation with contrastive learning. For state-based RL, \cite{sinha2022s4rl} propose several state-based transformation methods for the function approximation of the Q-networks. The adversarial data augmentations \cite{9528987,volpi2018generalizing,yin2023rerogcrl} generate the worst-case samples to augment the training process. Via iterative augmenting adversarial examples as the worst-case samples over data distributions that are near the source domain in the feature space, \cite{9528987} allows learning trained models that improve generalization across a prior unknown target domains. \cite{volpi2018generalizing} tackles the divergence-agnostic unsupervised domain adaptation problem, and they harness the data distribution and improve the generalization ability through the adversarial attack. None of them apply data augmentation in continual RL.

%% file: content/method.tex
\section{Data Augmentation for Continual RL}

In this section, we introduce how to use data augmentation for continual RL.
We start with the problem formulation in Section \ref{section:data_augmentation_framework}.
Then, in Section \ref{section:data_augmentation_methods}, we introduce 8 different data augmentation methods. Finally, we present the general data augmentation framework for continual RL in Section~\ref{section:data_augmentation_framework_of_continual_RL}.
\subsection{Problem Formulation} 
\label{section:data_augmentation_framework}
In the continual RL setting, there is a sequence of tasks. We assume that the boundary of the tasks can be accessed by the agent. Within a continual RL scenario, we consider a sequence of $N$ tasks, $S_N := (T_1, ... , T_N)$. Each task is referred to as the Markov Decision Process (MDP), consisting of a tuple of $<S^i,A^i,P^i,R^i,\gamma^i>$, where in task $i$, $S^i$ is the state space, $A^i$ is the action space, $P^i$ is the transition probability, $R^i: S^i \times A^i \to \mathbb{R} $ is the reward function to map the current state-action pair to a reward value, and $\gamma^i$ is the discount factor for mitigating the impact of long-term benefits. In continual RL, an agent needs to adapt its decision-making policy while encountering a sequence of tasks. The agent must be able to continually learn and adapt without losing information learned from past tasks. The ultimate objective of the agent is to optimize the policy parameters $\theta$ in order to maximize the average performance across all tasks: $\frac{1}{T_N}\sum_1^{T_N}{\mathbb{E}\sum_{k=0}^{\infty}{(\gamma^i)^k R^{i}_{t+k}}}$.

Data augmentation techniques serve as an effective method to enhance the diversity and quantity of training samples for each task. This is particularly crucial in continual learning processes, where an agent must interact with sequential environments. We denote the transformation as $\tilde{s}_t \sim \mathcal{T}(s_t)$, where $\tilde{s}_t$ represents the augmented state and $s_t$ the raw state. By incorporating the data augmentation $\mathcal{T}$ into the reinforcement learning (RL) update, we utilize the augmented sample $\tilde{s}_t$ to minimize the loss functions of both the policy and critic networks for each task.

\subsection{Data Augmentation Methods} \label{section:data_augmentation_methods}
In continual RL, since the agent endurances a sequence of tasks to collect new data samples and different tasks share the similar state space, data augmentation can allow the agent to generate more diverse data samples during each task. 
We describe the data augmentation methods as follows:

\noindent \textbf{Uniform noise} adds Uniform random variables $\textbf{z} \sim \mathcal{U}(-\alpha,\alpha)$ to raw state, $\tilde{s}_t=s_t + \textbf{z}$, where $\alpha$ is the bound of the Uniform noise.

\noindent \textbf{Gaussian noise} adds Gaussian random variables $\textbf{z} \sim \mathcal{N}(0,\sigma^2)$ to raw state, $\tilde{s}_t=s_t + \textbf{z}$, where $\sigma$ is the standard deviation of the Gaussian noise.

\noindent \textbf{Random amplitude scaling (RAS)} firstly introduced in \cite{laskin2020reinforcement}, $\tilde{s}_t=s_t \cdot \textbf{z}$, the original state $s_t$ scales with Uniform random variable $\textbf{z} \sim \mathcal{U}(\alpha,\beta)$. The goal of RAS is to scale the state values and to keep the signs of the states.

\noindent \textbf{Dimension dropout} randomly samples a dimension $d$ and make it equal to zero, $\tilde{s}_t=s_t \cdot  \textbf{z}$, where $\textbf{z} = [1,...,0,...,1],$ $d$ is sampled from Bernoulli Distribution. This transformation drops one dimension value of the current state.

\noindent \textbf{State-switch} flips the random selected dimensions within the state, while the intuition is that some dimensions retain the same physical meaning. When we transform the selected samples with different semantically properties, it will likely break the phisical assumption. In our setting, we flip the positions of two objects that the robot needs to manipulate. 

\noindent \textbf{Mixup} \cite{zhang2017mixup} is a well-used stochastic but easy-to-implement data augmentation technology. And it has also demonstrated effectiveness in RL \cite{wang2020improving}. We interpolate the current state and next state, $\tilde{s}_t = \lambda s_t + (1-\lambda) s_{t+1}$, where $\lambda \sim \mathrm{Beta}(\alpha, \beta)$. Based on \cite{zhang2017mixup}, we set the parameters of the beta distribution $\alpha = \beta$. To keep the stability of RL training, we keep the next state $s_{t+1}$ unchanged. 

\noindent \textbf{Adversarial Augmentation (Adv-AUG)} is to generate adversarial samples \cite{madry2017towards} and minimize the maximum loss over the adversarial samples. 


Formally, for each state point $s_i$ in replay buffer $\mathcal{D}$, we generate an adversarial data point $\tilde{s}_i$ that is similar to state $s_i$, and update the policy network $\pi_\theta$ by minimizing the maximum policy loss $\mathcal{L_{\pi_{\theta}}}(\tilde{s}_i)$ \cite{haarnoja2018soft} over $\tilde{s}_i$:
\begin{equation}
\underset{\pi_{\theta}}{\mathrm{min}} \; \underset{s_i \sim \mathcal{D}}{\mathbb{E}}[\underset{\tilde{s}_i : s_i + \delta}{\mathrm{max}}\mathcal{L_{\pi_{\theta}}}(\tilde{s}_i)], \quad \mathrm{s.t.} \quad ||\delta||_p<\epsilon
\end{equation}
where $\delta$ represents the perturbation distance, and $||\delta||_p <\epsilon$ means a $L_p$ norm ball centered at $s_i$ with radius $\epsilon$. The inner maximization is solved by multi-steps projected gradient decent and generating adversarial augmentation as below:
\begin{equation} \label{eq:adversarial_augmentation}
\tilde{s}_i \leftarrow \prod_{x+k}(s_i + \epsilon \cdot \mathrm{sign} \nabla_{s_i \sim \mathcal{D}}\mathcal{L}_{\pi_{\theta}}(s_i)),
\end{equation}
After obtaining the worst case for the policy update, we also use this adversarial sample $\tilde{s}_i$ to optimize the critic network, $\underset{Q_{\phi}}{\mathrm{min}} \; \underset{s_i \sim \mathcal{D}}{\mathbb{E}}[\underset{\tilde{s}_i : s_i + \delta}{\mathrm{max}}\mathcal{L}_{Q_{\phi}}(\tilde{s}_i)]$. 





\noindent \textbf{Adversarial Augmentation with GEM (Adv-GEM)}
Inspired by \cite{zhang2021generalization}, we propose a new method, Adv-GEM for continual RL. We provide details as follow.

While adversarial augmentation has proven very effective in the generalization and robustness of RL \cite{zhang2021generalization}, it cannot promise generalization performance by generating untargeted adversarial examples among several tasks in continual RL. Based on it, we propose adversarial augmentation with gradients episodic memory (Adv-GEM) for continual RL. 


\begin{figure}[th] 
\centering
\centerline{\includegraphics[width=0.8 \columnwidth]{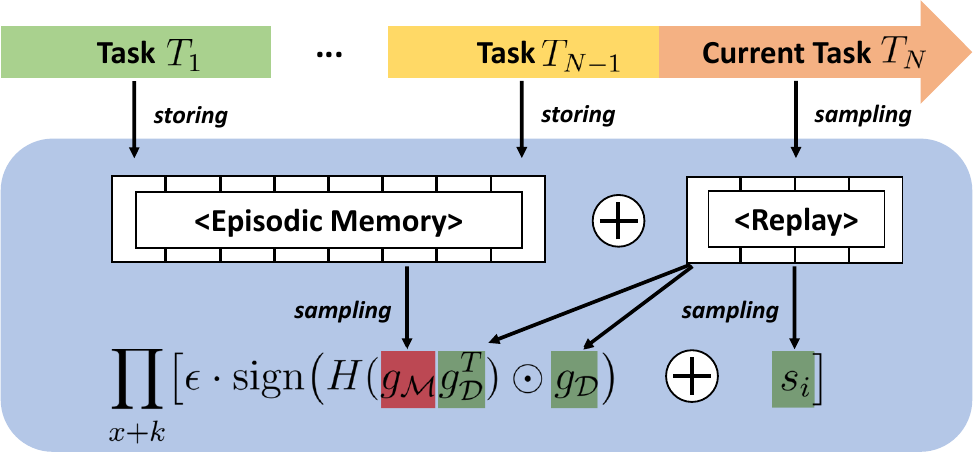}}
\caption{Framework for Adv-GEM data generation. 
}
\label{figure-adversarialDA}
\end{figure}

\input{content/alg_adv_gem.tex}

When we finish the past task $k$, we store a batch of samples in episodic memory $\mathcal{M}$. While we generate the adversarial samples $\tilde{s}_i$ on current task, we treat the average maximum worst cases $\underset{s_i \sim \mathcal{M}}{\mathbb{E}}[\underset{\tilde{s}_i : s_i + \delta}{\mathrm{max}}\mathcal{L}_{\pi_{\theta}}(\tilde{s}_i, \mathcal{M})] = \frac{1}{|\mathcal{\mathcal{M}}|}\sum_{s_i \in \mathcal{M}} \mathrm{max} \mathcal{L_{\pi_{\theta}}}(\tilde{s}_i)$ on the episodic memories of tasks $\mathcal{M}$, does not decrease. This effectively encourages the generated samples to be worse cases for previous tasks while minimizing the worst cases at the current task. 

According to adversarial augmentation, we choose PGD attack to generate the adversarial examples. To ensure that the episodic memory loss over the past tasks does not decrease, we compare the angle between the policy gradient vectors of previous tasks $g_\mathcal{M}$, and the gradient of current task $g_\mathcal{D}$. While the angle between $g_\mathcal{D}$ and $g_\mathcal{M}$ is greater than $90^{\circ}$, we project the proposed gradient to $0$. Otherwise, we will keep the gradient as $g_\mathcal{D}$. The corresponding adversarial examples can be calculated via:
\begin{equation} \label{eq:projected_adversarial_examples}
\tilde{s}_i \leftarrow \prod_{x+k}\bigl[s_i + \epsilon \cdot \mathrm{sign} \bigl(H(g_{\mathcal{M}}g_{\mathcal{D}}^{T})\odot g_{\mathcal{D}}\bigr)\bigr], \\
\end{equation}
\vskip -0.15in
\noindent where $H$ is the Heaviside step function, $g_{\mathcal{D}}$ is the gradient of current task, $g_{\mathcal{D}} = \nabla_{s_i \sim \mathcal{D}}\mathcal{L}_{\pi_{\theta}}(s_i)$, and $g_{\mathcal{M}}$ is the gradient of previous task, $g_{\mathcal{M}} = \nabla_{s_i \sim \mathcal{M}}\mathcal{L}_{\pi_{\theta}}(s_i)$. Intuitively, when the gradient $g_\mathcal{D}$ violates the constraint, it is projected to 0. Therefore,  we only keep the gradient which of the angle between the gradient of the previous task. Figure \ref{figure-adversarialDA} demonstrates the Adv-GEM during the continual RL learning process. 

Algorithm~\ref{alg_1} summarizes the generation of Adv-GEM examples. Adv-GEM starts with data collection by interacting with the environment and then adds the transitions $\{(s_{\mathcal{D}_i}, a_{\mathcal{D}_i}, r_{\mathcal{D}_i}, s_{\mathcal{D}_i}')\}$  to the episodic replay buffer $\mathcal{M}$. 
Via Adv-GEM augmentation, we obtain the augmentation samples according to Eq.~\ref{eq:projected_adversarial_examples}. Finally, at the end of the task, we store $N_\mathcal{M}$ samples in the episodic memory $\mathcal{M}$. To verify the impact of the episodic memory size, we perform the ablation study at section~\ref{ablation}. 
Via embedding the gradient of episodic memory into data augmentation and generating the adversarial examples, Adv-GEM can mitigate the inherent overfitting problem of the memory-based methods \cite{lopez2017gradient,chaudhry2018efficient}. Adv-GEM offers several advantages in continual learning. By incorporating gradients from previous tasks, it encourages learning features useful across multiple tasks, leading to more generalized representations. This addresses the stability-plasticity dilemma by balancing adaptation to new information with retention of past knowledge. Adv-GEM's adversarial examples expand the training distribution relevantly, guiding exploration towards robust solutions. By aligning generated examples with previous task gradients, it implicitly regularizes learning, reducing catastrophic forgetting. These insights suggest Adv-GEM produces robust, generalizable representations that maintain performance across tasks while avoiding overfitting, addressing a key challenge in continual RL.


\input{content/alg_DA}

\subsection{Data Augmentation Framework of Continual RL} \label{section:data_augmentation_framework_of_continual_RL}
For continual RL, we use Soft Actor-Critic (SAC) \cite{haarnoja2018soft}, an off-policy algorithm with entropy regularization, as the backbone algorithm. Given data augmentation $\mathcal{T}$ and original transitions $\{(s_{\mathcal{D}_i}, a_{\mathcal{D}_i}, r_{\mathcal{D}_i}, s_{\mathcal{D}_i}')\}$ from replay buffer $\mathcal{D}$, we follow a similar scheme as DrQ \cite{kostrikov2020image}, where we average the policy value, Q value, and target Q value over augmented states and original states. Therefore we have more diverse data for optimization of policy network and critic network.


We deploy the data augmentation scheme on the continual RL. The overall algorithm is depicted in Algorithm~\ref{alg:data_augmentation}. For each task, the RL agent starts with data collection by interacting with the environment and then adds the transitions to the replay buffer $\mathcal{D}$. At each update step for policy and critic networks, we sample mini-batch from the replay buffer $\mathcal{D}$. Given transformation $\mathcal{T}$, we average the Q value, target Q value and policy value among original states and augmented states in a mini-batch of transitions $\{(s_i, a_i, r_i, s_i^{\prime})\}_{i=1}^{\mathcal{B}_{\mathcal{D}}}$. The loss of the
policy network is optimized by the averaged policy value and Q value. Then we use the averaged target Q value and Q value in the Bellman equation to update the Q network. By generating such augmented data samples, we are able to learn an agent with better generalization ability for the continual learning process.

In continual RL, an agent must adapt to a sequence of tasks, each requiring interaction with a different environment to collect new data samples. Data augmentation serves as a powerful yet straightforward technique to enhance the diversity of data for each task. By applying augmentation to state representations, we expand the range of physically realizable inputs within each task's domain. Given that the state and action spaces remain consistent across tasks in continual RL, data augmentation increases the likelihood that samples from future tasks will align more closely with the current task's data distribution. This alignment facilitates better generalization and knowledge transfer between tasks, potentially mitigating catastrophic forgetting and improving overall performance in continual learning scenarios.

%% file: content/alg_adv_gem.tex
\begin{algorithm}[ht] 
    \caption{Adv-GEM Augmentation}
    \label{alg_1}
\begin{algorithmic}[1]
\State{\textbf{Input:}  replay buffer each task  $\mathcal{D}$, episodic memory buffer $\mathcal{M}$, episodic memory size each task $N_\mathcal{M}$}
\For {each continual task}
    \State $\{(s_{\mathcal{D}_i}, a_{\mathcal{D}_i}, r_{\mathcal{D}_i}, s_{\mathcal{D}_i}')\} \sim\mathcal{D}$
    \State $\{(s_{\mathcal{M}_i}, a_{\mathcal{M}_i}, r_{\mathcal{M}_i}, s_{\mathcal{M}_i}')\} \sim \mathcal{M}$
    \State {$\tilde{s}_i \gets$ $\mathcal{T}_\text{Adv-GEM}(s_{\mathcal{D}_i}, s_{\mathcal{M}_i})$  by Eq.\ref{eq:projected_adversarial_examples}} \label{alg:aug_method}
\EndFor
\State $\mathcal{M} \gets \mathcal{M} \cup \underset{size=N_\mathcal{M}}{\operatorname{RandomSample}}(\mathcal{D})$ 
\end{algorithmic}
\end{algorithm}

%% file: content/alg_DA.tex
\begin{algorithm}[tb] 
    \caption{Data Augmentation for Continual RL}
    \label{alg:data_augmentation}
\begin{algorithmic}[1]
    \State \textbf{Hyperparameters:} Policy Network parameter $\theta$, Critic Network parameter $\phi$, time steps per task $T$, policy learning rate $\eta$, critic learning rate $\beta$, replay buffer each task  $\mathcal{D}$, batch size of the replay buffer $\mathcal{B}_\mathcal{D}$, data augmentation $\mathcal{T}$, 
    \State \textbf{Initialization:} Policy Network $\theta_0$, Critic Network $\phi_0$
    \For {each continual task}
        \For {each timestep $t = 1 \, \mathrm{to} \, T$}
            \State{$a_t \sim \pi(\cdot \mid s_t)$}
            \State{${s_t}' \sim P(\cdot\mid s_t, a_t)$}
            \State {$\mathcal{D} \gets \mathcal{D} \cup (s_t, a_t, r{(s_t, a_t)}, s_t')$}
            \State $\{(s_i, a_i, r_i, s_i^{\prime})\}_{i=1}^{\mathcal{B}_{\mathcal{D}}} \sim\mathcal{D}$
            \State $Q_{\phi} = \mathbb{E} [Q_{\phi}(s_i, a_i) + Q_{\phi}(\mathcal{T}(s_i), a_i)]$
            \State $V_{\phi} = \mathbb{E} [V_{\phi}(s^{\prime}_i) + V_{\phi}(\mathcal{T}(s^{\prime}_i))]$
            \State $\pi_{\theta} = \mathbb{E}[\pi_{\theta}(\cdot \mid s_t) + \pi_{\theta}(\cdot\mid \mathcal{T}(s_t))]$
            \State $\theta \leftarrow \theta - \eta  \nabla [\mathrm{log}\pi_{\theta} - Q_{\phi}]$
            \State $\phi \leftarrow \phi - \beta  \nabla [Q_{\phi} - (r + \gamma V_{\phi})]^2$
        \EndFor
    \EndFor
\end{algorithmic}

\end{algorithm}

%% file: content/experiment.tex
\section{Experiments}
To evaluate the effectiveness of our data augmentation methods, we compare a variety of approaches. We first describe the environments,  methods, and evaluation metrics for continual RL in Section~\ref{experiment_setting}. In Section~\ref{section:comparison_to_the_continual_rl_methods}, we compare the various baseline on continual experiment settings. We provide the results of ablation study in Section~\ref{ablation}. 

\begin{figure}[th] 
\centering
\centerline{\includegraphics[width= \columnwidth]{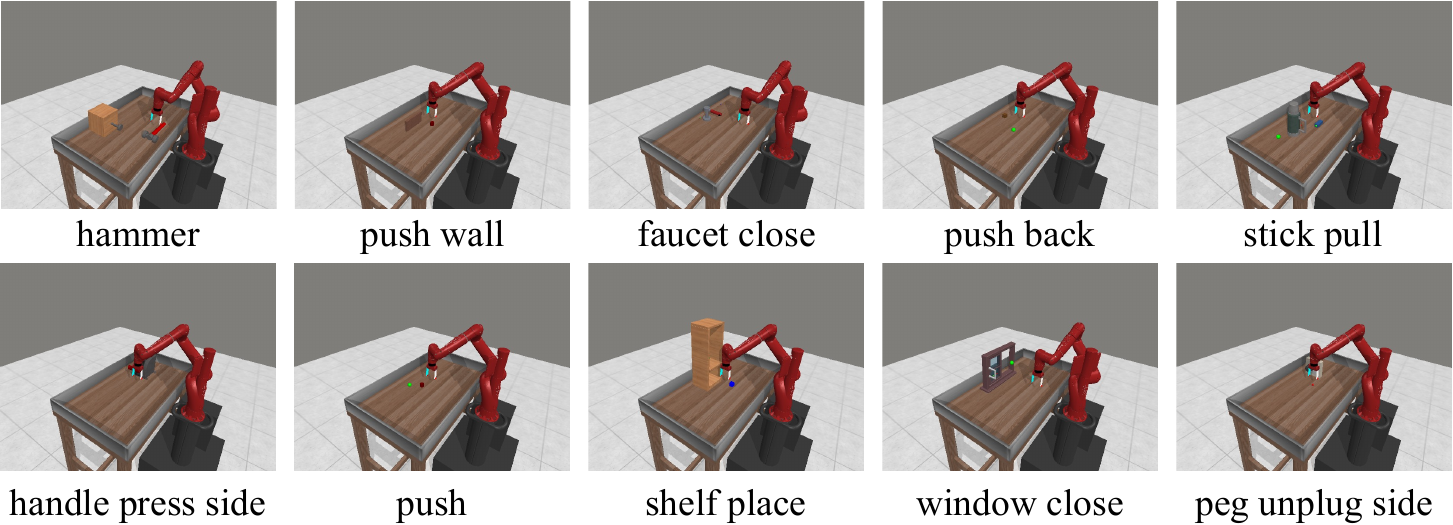}}
\caption{CW10 consists of 10 manipulation tasks, carefully designed to be diverse while maintaining a shared structure. This shared structure facilitates efficient continual RL.}
\label{figure-cw10}
\end{figure}

\subsection{Experiment Setting} \label{experiment_setting}
\textbf{Environments.} To efficiently evaluate models, we build a sequence of tasks, dubbed \textbf{MW4}, which contains 4 manipulator tasks from Meta World \cite{yu2020meta}.
Specifically, we consider tasks including \textit{hammer}, \textit{push  wall}, \textit{faucet close}, and \textit{push back}, which share the same state and action spaces with 12-dimensional vectors and 4-dimensional vectors respectively. 
In addition, we also consider using a larger set of tasks and we choose \textbf{CW10} from Continual World \cite{wolczyk2021continual} as shown in Fig.~\ref{figure-cw10}, which contains 10 sequential tasks in total. The CW10 sequence contains \textit{hammer, push wall, faucet close, push back, stick pull, handle press side, push, shelf place, window close, peg unplug side}.

\noindent \textbf{Methods.} For continual RL, to conduct the comprehensive comparison, we consider using the following state-of-the-art continual RL methods, including one regularization-based method, EWC \cite{kirkpatrick2017overcoming},
and one modality-based method, PackNet \cite{mallya2018packnet}. 

For data augmentation, we combine our proposed methods with EWC and PackNet to show the effectiveness. We name the combination methods EWC + Uniform, EWC + Gaussian, EWC + RAS, EWC + dimension-dropout, EWC + state-switch, EWC + mixup, EWC + Adv-AUG, EWC + Adv-GEM, etc. Our proposed data augmentation methods are orthogonal to existing continual RL methods. Thus, they can be seamlessly combined.
With limited computing resources, we only demonstrate the result of EWC and PackNet. Combining with other continual RL methods is straightforward.

\noindent \textbf{Evaluation Metrics.} \label{evaluation-metrics} 
We report the average performance, forward transfer and catastrophic forgetting as below.
\begin{itemize}[leftmargin=*,label=\textbullet]
\item \textbf{Average Performance.} The performance of task $i$ is represented by its success rate, $p_i(t) \in [0,1]$ at time $t$. The average performance of all tasks at time $t$ is described as:
\begin{equation}
\mathrm{P}(t) := \frac{1}{N}\sum_{i=1}^{N}{p_i(t)}.\nonumber
\end{equation}
\item \textbf{Forward Transfer.}
The forward transfer is measured by the normalized area between its training curve and the training curve of the single task performance. The $\mathrm{FT}$ is the average Forward Transfer at all tasks: 
\begin{equation}
\mathrm{FT}:=\frac{1}{N}\sum_{i=1}^{N}\frac{\frac{1}{\Delta}\int_{(i-1)\cdot\Delta}^{i\cdot\Delta}p_i(t)dt - \frac{1}{\Delta}\int_{(i-1)\cdot\Delta}^{i\cdot\Delta}p_i^b(t)dt}{1-\frac{1}{\Delta}\int_{(i-1)\cdot\Delta}^{i\cdot\Delta}p_i^b(t)dt}.\nonumber
\end{equation}

\item \textbf{Catastrophic Forgetting.} For each task $i$, we measure the performance decrease after the whole training process for each tasks, by comparing with the performance of the end of task $i$, defined as
\begin{equation}
\mathrm{CF} := \frac{1}{N}\sum_{i=1}^{N}{(p_i(i \cdot \Delta) - p_i(T))}.\nonumber
\end{equation}
\end{itemize}

\subsection{Results} \label{section:comparison_to_the_continual_rl_methods}

In this section, we will first investigate different data augmentation methods on the MW4 experiment to figure out which augmentation is more helpful for continual RL. To validate the effectiveness, we add data augmentation methods described above to EWC and PackNet at MW4 environments. Additionally, in order to verify the longer task scenarios, we conduct the experiment at CW10 with 10 robots control tasks in total. 

\begin{figure}[H]
\centering
\centerline{\includegraphics[width=0.99\linewidth]{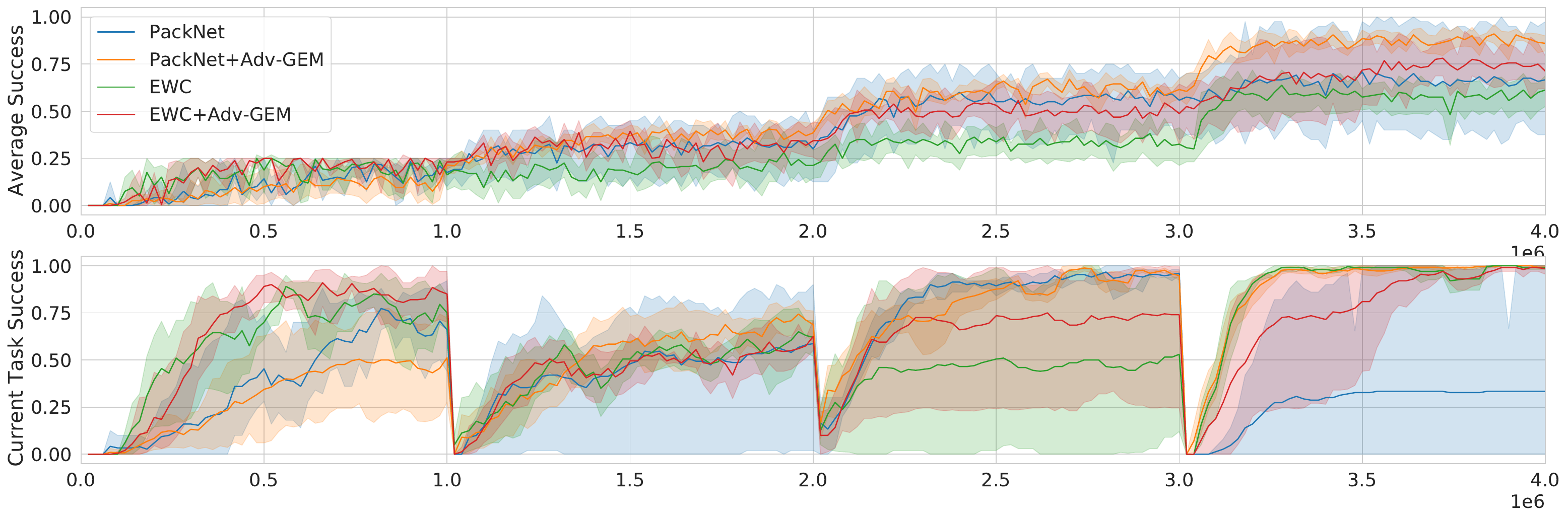}}
\caption{(Top) Average Performance in MW4 for EWC and PackNet with Adv-GEM. (Bottom) The test success rate on the current task in MW4 for EWC and PackNet. 
}
\label{figure:AveragePerformance_EWC_PackNet}
\end{figure}

\begin{table}[t]
\caption{Comparison to continual RL baselines on MW4 by combining our proposed data augmentation methods with EWC and PackNet. The results are 90\% confidence intervals based on runs of 5 seeds. 
}
\label{table:Results_on_CW4}
\centering
\resizebox{0.9\linewidth}{!}{
\input{content/table_MW4.tex}
}
\vspace{-10pt}
\end{table}

\noindent \textbf{Effect of Data Augmentation.} To validate the effect of data augmentation at MW4, we compare a base EWC agent and a base Packnet agent with various data augmentation methods. The average performance, forgetting, and forward transfer are tabulated in Table.~\ref{table:Results_on_CW4}. We found that RAS, state-switch, mixup, Adv-AUG, and Adv-GEM outperform the base methods on average performance and forward transfer both at EWC and PackNet. For the average performance, the EWC + Adv-GEM and PackNet + Adv-GEM methods are 0.74 and 0.88 respectively, suggesting their effectiveness over different continual RL methods. However, we also see that Uniform noise, Gaussian noise, and dimension-dropout can hurt the average performance of the continual RL agent when it is combined with EWC. For PackNet method, the Uniform noise and dimension-dropout augmentation decreased the average performance. These declined performances may be because they generate out-of-domain augmented data. The forward transfer represents the ability that continual methods to transfer knowledge forward. We found that RAS, mixup, Adv-AUG and Adv-GEM can highly increase the forward transfer both with EWC and PackNet. 
EWC + Adv-GEM and PackNet + Adv-GEM reduce catastrophic forgetting to 0.06 and -0.10 respectively, demonstrating data augmentation's ability to enhance data distribution diversity.

Figure~\ref{figure:AveragePerformance_EWC_PackNet} demonstrates that Adv-GEM augmentation significantly improves the average performance of both EWC and PackNet. EWC + Adv-GEM even outperforms PackNet, due to better test performance in tasks 1 and 3. PackNet + Adv-GEM surpasses PackNet with higher performance in task 4. Overall, these findings not only demonstrate the effectiveness of data augmentation in continual RL but also highlight the potential of Adv-GEM as a promising approach to addressing key challenges in the field, such as catastrophic forgetting and efficient knowledge transfer between tasks.




\begin{table}[t]
\caption{Comparison to continual RL baselines on CW10 by combining our proposed data augmentation methods with EWC and PackNet. The results are run for 3 seeds with 90\% confidence intervals. 
}
\label{table:Results_on_CW10}
\centering
\resizebox{0.9\linewidth}{!}{
\input{content/table_CW10.tex}
}
\end{table}

\begin{wraptable}{R}{0.4\linewidth}
\vspace{-30pt}
{
\caption{Average performance of different memory sizes for MW4.}
\label{table:Memory_Impact_Average_Performance}
\centering
\resizebox{\linewidth}{!}{
\input{content/table_memory}}
}
\vspace{-10pt}
\end{wraptable}

\noindent \textbf{Experiments at CW10.} To assess the effectiveness of longer task scenarios, we evaluate the impact of data augmentations on CW10 environments. We apply all augmentation methods at the EWC and PackNet methods. As shown in Table~\ref{table:Results_on_CW10}, RAS, Adv-AUG, and Adv-GEM improve the average performance on CW10 when combined with EWC. Notably, the Adv-GEM transformation achieves a higher average performance value than the Adv-AUG transformation. In terms of catastrophic forgetting and forward transfer, both Adv-AUG and Adv-GEM perform similarly to the original EWC method. As for PackNet, both Adv-AUG and Adv-GEM outperformed the original method in terms of average performance. The Adv-AUG + PackNet has 0.3 forward transfer value. These results showcase that data augmentations, like RAS, state-switch, mixup, Adv-AUG, Adv-GEM can improve the average performance of EWC and PackNet at MW4. The consistent improvement across EWC and PackNet algorithms, and MW4 and CW10 benchmarks, highlights the versatility and robustness of these data augmentation techniques. This suggests they address fundamental continual learning challenges like catastrophic forgetting and knowledge transfer, rather than exploiting specific algorithm or environment characteristics.


\noindent \textbf{Effects of memory buffer size.} To examine the impact of episodic memory size on model performance in MW4, we evaluated EWC and PackNet with memory sizes of 0, 5k, and 10k. Table.~\ref{table:Memory_Impact_Average_Performance} presents these results. The Adv-GEM method consistently outperforms the baseline across various memory buffer sizes. Notably, with a memory size of 10k, both EWC + Adv-GEM and PackNet + Adv-GEM achieve the highest average performance. Considering the memory constraints typical in continual RL, we selected 10k as our standard memory size for subsequent experiments.

\input{content/ablation_study}

%% file: content/table_MW4.tex
\begin{tabular}{lccc}
\toprule
\textbf{Method} & \textbf{performance $\uparrow$} & \textbf{forgetting $\downarrow$} & \textbf{f. transfer $\uparrow$} \\
\midrule
EWC & 0.60 \scriptsize{[0.53, 0.66]}& 0.13 \scriptsize{[0.03, 0.20]}& -0.04 \scriptsize{[-0.36, 0.29]} \\
EWC + Uniform & 0.58 \scriptsize{[0.46, 0.70]}& 0.09 \scriptsize{[ 0.03, 0.15]}& -0.23 \scriptsize{[-0.48, 0.02]} \\
EWC + Gaussian &  0.45 \scriptsize{[0.31, 0.58]}& 0.13 \scriptsize{[ 0.09, 0.18]}& -0.33  \scriptsize{[ -0.66, 0.00]} \\
EWC + RAS & 0.70 \scriptsize{[0.66, 0.74]} & 0.10 \scriptsize{[0.02, 0.17]} & \textbf{0.14  \scriptsize{[0.00, 0.28]}} \\
EWC + dimension-dropout & 0.53 \scriptsize{[0.47, 0.58]}& 0.14 \scriptsize{[0.05, 0.22]}& -0.19 \scriptsize{[-0.48, 0.11]} \\
EWC + state-switch &  0.68 \scriptsize{[0.60, 0.75]} & 0.14 \scriptsize{[0.05, 0.21]}& 0.12 \scriptsize{[-0.00, 0.24]} \\
EWC + mixup & 0.67 \scriptsize{[0.62, 0.72]} & \textbf{0.03 \scriptsize{[-0.05, 0.12]}} & 0.03 \scriptsize{[-0.19, 0.24]} \\
EWC + Adv-AUG & 0.66 \scriptsize{[0.54, 0.76]} & 0.14 \scriptsize{[0.06, 0.20]} & 0.12 \scriptsize{[-0.06, 0.29]} \\
EWC + Adv-GEM & \textbf{0.74 \scriptsize{[0.65, 0.84]}}& 0.06 \scriptsize{[-0.07, 0.20]} & 0.09 \scriptsize{[-0.15, 0.32]} \\ \midrule
PackNet  & 0.73 \scriptsize{[0.52, 0.94]}& -0.03 \scriptsize{[-0.07, 0.00]}& -0.03 \scriptsize{[-0.39, 0.34]} \\
PackNet + Uniform & 0.68 \scriptsize{[0.56, 0.79]} & 0.03 \scriptsize{[-0.02, 0.09]} & -0.19 \scriptsize{[-0.44, -0.01]}\\
PackNet + Gaussian & 0.78 \scriptsize{[0.70, 0.85]} & -0.07 \scriptsize{[-0.08, -0.06]} & -0.19 \scriptsize{[-0.39, 0.02]}\\
PackNet + RAS & 0.87 \scriptsize{[0.84, 0.90]} & -0.01 \scriptsize{[-0.05, 0.03]} & 0.10 \scriptsize{[-0.08, 0.27]} \\
PackNet + dimension-dropout & 0.54 \scriptsize{[0.44, 0.65]} & 0.01 \scriptsize{[-0.03, 0.05]} & -0.41 \scriptsize{[-0.71, -0.10]}\\
PackNet + state-switch & 0.81 \scriptsize{[0.73, 0.89]} & 0.00 \scriptsize{[-0.04, 0.04]} & 0.06 \scriptsize{[-0.11, 0.18]} \\
PackNet + mixup& 0.81 \scriptsize{[0.77, 0.86]} & 0.03 \scriptsize{[-0.01, 0.06]} & \textbf{0.16 \scriptsize{[0.04, 0.27]}} \\
PackNet + Adv-AUG & 0.82 \scriptsize{[0.77, 0.87]} & -0.02 \scriptsize{[-0.04, 0.01]} & 0.04 \scriptsize{[-0.13, 0.20]} \\
PackNet + Adv-GEM & \textbf{0.88 \scriptsize{[0.86, 0.90]}} & \textbf{-0.10 \scriptsize{[-0.14, -0.05]}} & 0.10 \scriptsize{[-0.02, 0.22]} \\
\bottomrule
\end{tabular}

%% file: content/table_CW10.tex
\begin{tabular}{lccc}
\toprule
\textbf{Method.} & \textbf{performance $\uparrow$} & \textbf{forgetting $\downarrow$} & \textbf{f. transfer $\uparrow$} \\
\midrule
EWC & 0.66 \scriptsize{[0.62, 0.69]} & 0.03 \scriptsize{[0.01, 0.06]} & 0.05 \scriptsize{[-0.02, 0.12]} \\
EWC + Uniform & 0.01 \scriptsize{[0.00, 0.01]}&  \textbf{-0.01 \scriptsize{[-0.01, 0.00]}}&  -1.94 \scriptsize{[-2.24, -1.63]} \\
EWC + Gaussian & 0.02 \scriptsize{[0.00, 0.03]}&  0.00 \scriptsize{[-0.01, 0.02]}& -1.86 \scriptsize{[-2.15, -1.57]} \\
EWC + RAS & 0.67 \scriptsize{[0.66, 0.68]} &  0.00 \scriptsize{[-0.01, 0.00]}& \textbf{0.06 \scriptsize{[-0.01, 0.11]}} \\
EWC + dimension-dropout & 0.37 \scriptsize{[0.35, 0.39]}& 0.02 \scriptsize{[0.01, 0.02]}&  -1.00 \scriptsize{[-1.25, -0.79]} \\
EWC + state-switch & 0.66 \scriptsize{[0.61, 0.71]}& \textbf{-0.01 \scriptsize{[-0.02, -0.01]}}& \textbf{0.06 \scriptsize{[-0.01, 0.12]}} \\
EWC + mixup & 0.58 \scriptsize{[0.52, 0.63]}& 0.07 \scriptsize{[0.03, 0.11]}& -0.19 \scriptsize{[-0.43, 0.06]} \\
EWC + Adv-AUG & 0.67 \scriptsize{[0.55, 0.79]} & 0.06 \scriptsize{[0.03, 0.09]} & -0.03 \scriptsize{[-0.12, 0.06]}  \\
EWC + Adv-GEM & \textbf{0.72 \scriptsize{[0.60, 0.84]}}& 0.05 \scriptsize{[0.01, 0.09]} & 0.04 \scriptsize{[-0.08, 0.16]} \\ 
\midrule
PackNet & 0.83 \scriptsize{[0.81, 0.85]} & -0.00 \scriptsize{[-0.01, 0.01]} & 0.21 
\scriptsize{[0.16, 0.25]} \\
PackNet + Uniform & 0.59 \scriptsize{[0.53, 0.65]} & -0.01 \scriptsize{[-0.03, 0.02]} & -0.54 \scriptsize{[-0.86, -0.22]}\\
PackNet + Gaussian & 0.70 \scriptsize{[0.67, 0.73]} & 0.01 \scriptsize{[-0.00, 0.02]} & -0.03 \scriptsize{[-0.08, 0.03]}\\
PackNet + RAS & 0.79 \scriptsize{[0.72, 0.85]} & -0.03 \scriptsize{[-0.03, -0.02]} & 0.00 \scriptsize{[-0.24, 0.24]} \\
PackNet + dimension-dropout & 0.46 \scriptsize{[0.40, 0.52]} & 0.09 \scriptsize{[0.07, 0.10]} & -0.56 \scriptsize{[-0.94, -0.17]}\\
PackNet + state-switch & 0.88 \scriptsize{[0.83, 0.93]} & \textbf{-0.10 \scriptsize{[-0.11, -0.08]}} & 0.07 \scriptsize{[-0.19, 0.33]} \\
PackNet + mixup& 0.83 \scriptsize{[0.82, 0.84]} & -0.04 \scriptsize{[-0.05, 0.04]} & 0.10 \scriptsize{[-0.03, 0.22]} \\
PackNet + Adv-AUG & \textbf{0.89 \scriptsize{[0.86, 0.92]}} & -0.01 \scriptsize{[-0.01, -0.01]} & \textbf{0.30 \scriptsize{[0.26, 0.33]}}\\
PackNet + Adv-GEM & 0.88 \scriptsize{[0.84, 0.91]} & -0.01 \scriptsize{[-0.02, -0.01]} & 0.19 \scriptsize{[0.09, 0.29]} \\
\bottomrule
\end{tabular}

%% file: content/table_memory.tex
\begin{tabular}{@{}cccc@{}}
\toprule
\multicolumn{4}{c}{\textbf{MW4}}        \\ \midrule
\textbf{Memory Size} & 0 & 5k & 10k \\ \midrule
EWC + Adv-GEM        & 0.66  &  0.73    & \textbf{0.74 } \\ \midrule
PackNet + Adv-GEM   & 0.82 & 0.85 & \textbf{0.88}   \\ \bottomrule
\end{tabular}

%% file: content/ablation_study.tex
\label{ablation}

\begin{wraptable}{r}{0.4\linewidth}
\vspace{-30pt}
{
\caption{Average performance of different epsilons for MW4.}
\label{table:Epsilon_Impact_Average_Performance}
\resizebox{\linewidth}{!}{
\input{content/table_epsilon}}
}
\vspace{-5pt}
\end{wraptable}
\noindent \textbf{Epsilon.} To investigate the effect of different epsilon values, which is the adversarial attack size, in Adv-AUG and Adv-GEM, we compare the value of 0.1, and 0.01 in MW4 respectively. As the scale of the input states, the epsilon value should be selected properly. Table.~\ref{table:Epsilon_Impact_Average_Performance} shows that $\epsilon = 0.1$ has better average performance. Due to the limitation of the computing source, we did not increase the epsilon to test. According to the scale of state input at the CW10 experiment, there is no need to increase the adversarial attack size over 1, which would exceed the physical constraints of the system.

\noindent \textbf{Computation efficiency.} Table.~\ref{table:Computation_efficiency} shows the efficiency comparison results. We ran our experiments on Ubuntu 18.04 with Nvidia A100 GPU
and 40G RAM. We set the simple baseline EWC and PackNet with a running time unit of 1. Most of our data augmentation methods are efficient enough, and their running time of them is around 0.8-1.5 times compared with the baseline methods. Our EWC + Adv-AUG method has 2.4 times computing cost compared with EWC methods, the reason is that the adversarial attack needs to calculate the gradient of the policy network. Our Adv-GEM method has 3-4 times the computational cost compared to either EWC or PackNet baseline. In future work, we will improve our Adv-GEM method's computational efficiency.

\begin{table}[h]
\centering
\vspace{-10pt}
{\caption{Comparison of the computation efficiency between data augmentation methods and the baseline methods. 
}
\label{table:Computation_efficiency}
\input{content/table_computation}}
\end{table}

%% file: content/table_epsilon.tex
\begin{tabular}{@{}cccc@{}}
\toprule
\multicolumn{3}{c}{\textbf{MW4}}        \\ \midrule
\textbf{Epsilon} & 0.1 & 0.01 \\ \midrule
EWC + Adv-AUG  &  \textbf{0.66} & 0.61 \\ 
EWC + Adv-GEM &  \textbf{0.74} & 0.70  \\ \midrule
PackNet + Adv-AUG  &  \textbf{0.82} & 0.75 \\ 
PackNet + Adv-GEM   & \textbf{0.88}  & 0.80  \\ \bottomrule
\end{tabular}

%% file: content/table_computation.tex
\begin{tabular}{@{}cccc@{}}
\toprule
\multicolumn{4}{c}{\textbf{MW4}}        \\ \midrule
\textbf{Algorithm} & RunTime & \textbf{Algorithm} & RunTime  \\ \midrule
EWC                  & 1.0 &PackNet              & 1.0    \\
EWC + Uniform & 1.4 & PackNet + Uniform & 0.9\\
EWC + Gaussian & 1.4 & PackNet + Gaussian & 0.8\\
EWC + RAS & 1.1 & PackNet + RAS & 1.5\\
EWC + dimension-dropout & 1.2 & PackNet + dimension-dropout & 1.0\\
EWC + state-switch & 1.1 &PackNet + state-switch & 1.0\\
EWC + mixup & 1.1 &PackNet + mixup & 1.1\\
EWC + Adv-AUG  &  2.4 &PackNet + Adv-AUG  & 2.2\\
EWC + Adv-GEM &  3.5 &PackNet + Adv-GEM &   3.4\\ 
\bottomrule
\end{tabular}

%% file: content/conclusion.tex
\vspace{-20pt}
\section{Conclusion}

This paper introduces a data augmentation benchmark for state-based continual RL, showing significant improvements in performance and forward transfer. We present adversarial augmentation with memory gradient, enhancing data diversity through gradient episodic memory. Our flexible framework is easily expandable. Future work will focus on improving augmentation efficiency, validating the approach in diverse real-world applications, and developing adaptive strategies for varying task complexities.